\documentclass{sig-alternate-05-2015}

\usepackage{tikz}

\begin{document}

\setcopyright{acmcopyright}



\conferenceinfo{ODD workshop, SIGKDD '18}{August 20, 2018, London, GB}


%

\title{Are generative deep models for novelty detection truly better?
}
%
%
%
%
%

\numberofauthors{3} 
%
\author{
%
%
\alignauthor
V\'it \v{S}kv\'ara\\
       \affaddr{UTIA CAS CR}\\
       \affaddr{Pod Vod\'arenskou v\v{e}\v{z}\'i 4}\\
       \affaddr{Prague, Czech Republic}\\
       \email{skvara@utia.cas.cz}
\alignauthor
Tom\'a\v{s} Pevn\'y \\
       \affaddr{FEL CTU Prague}\\
       \affaddr{Technick\'a 2}\\
       \affaddr{Prague, Czech Republic}\\
       \email{pevnytom@fel.cvut.cz}
\alignauthor 
V\'aclav \v{S}m\'idl\\
       \affaddr{UTIA CAS CR}\\
       \affaddr{Pod Vod\'arenskou v\v{e}\v{z}\'i 4}\\
       \affaddr{Prague, Czech Republic}\\
       \email{smidl@utia.cas.cz}
}

\maketitle
\begin{abstract}
Many deep models have been recently proposed for anomaly detection. This paper presents comparison of selected generative deep models and  classical anomaly detection methods on an extensive number of non--image benchmark datasets. We provide  statistical comparison of the selected models, in many configurations, architectures and hyperparamaters.  We arrive to conclusion that performance of the generative models is determined by the process of selection of their hyperparameters. Specifically, performance of the deep generative models deteriorates with decreasing amount of anomalous samples used in hyperparameter selection. In practical scenarios of anomaly detection, none of the deep generative models systematically outperforms the kNN.
\end{abstract}

%
%


%
%

%
%



\section{Introduction}
  An \text{anomaly} is a data sample that is so different from the rest of the \text{normal} data that it was likely generated by a different underlying process. Anomalous samples are object of interest for various reasons and methods for novelty or anomaly detection try to identify them. These methods are used in a plethora of domains, including medical, computer security or sensor data collection. An overview of anomaly detection methods is presented in \cite{pimentel2014review}.

  This paper asks the following question -- how do anomaly detection methods based on deep neural--network generative models stand in comparison to the methods based on alternative paradigms? In the fashion of \cite{odena2018realistic} or \cite{campos2016evaluation}, we do not propose a new algorithm to prove it performs better than the existing methods. Instead, we aim to clarify whether the existing generative models actually bring a significant improvement over classical approaches. To our knowledge, such comparison of anomaly detection methods based on neural networks to some simpler methods is missing. Sadly, most papers proposing new methods (especially those based on deep neural networks)  focuse on large image data from MNIST, CIFAR or some other publicly available database that pose an unsurmountable task for classical anomaly detection methods. However, image--based anomaly problems are a rather niche domain, therefore this paper focuses on evaluation on a varied range of datasets. When presenting a novel generative model, the authors usually compare to some baseline generative model against which they are trying to improve, a limited number of classical models such as PCA or OSVM, and at best they use the KDD dataset -- see e.g. \cite{an2015variational} or \cite{xu2018unsupervised}.

 Furthermore, no one has ever done a thorough comparative study of generative deep models on a large number of different datasets. In the following text, a handful of selected generative models will be compared against each other and classical novelty detection methods in a statistical way on a large number of carefully crafted benchmark datasets. We do not claim to provide a complete overview of generative models applicable to anomaly detection task, but a general comparison that may simplify future assessment of novel methods. We admit that we have not tested or implemented state--of--the art methods such as \cite{zong2018deep} or \cite{zhai2016deep}.
 
 Additionally, we endeavor to create a standardized, publicly available implementation of different models. Apart from the actual model implementations, we create a framework for proper training, testing and comparing the models. The framework provides an interface for easy evaluation of new algorithms.
 
 Finally, we propose a number of important questions that may lead to a better understanding of individual tested algorithms and their behavior on different datasets. Using experimental data and a thorough comparison methodology, we may be able to answer some of these questions or provide valuable insight.



\section{Background}
Generative models are used to generate samples from some learned data distribution $p(x)$. In the case of anomaly detection, this is the distribution of normal data. Therefore, even though all following methods are unsupervised by default, we strive to train the generative model with (mostly) normal data. Even though it may be expensive to obtain labels, they are also useful for tuning of hyperparameters, as will be discussed later.

When the model has learned the normal data distribution, it can be used to compute an anomaly score function $f:\mathcal{X} \rightarrow \mathbb{R}$ for a sample $x \in \mathcal{X}$ from the data space. The convention used here is that higher the anomaly score, the more likely it is that the sample is an anomaly. For generative models, reconstruction error (in case of autoencoders), discriminator score (for adversarial models) or their combination can be used. 

This section contains a brief theoretical background of the used generative models based on neural nets and two classical methods. The description of cost functions, anomaly score functions and their parameters will be given.

\subsection{Autoencoding models}
\subsubsection{Autoencoder}
  An autoencoder (AE) is a cornerstone of many multi layer perceptron (MLP) models. Although not a generative model by itself, it may be used as a baseline for the variational autoencoder generative model. It is easily used for the anomaly detection task (see \cite{sakurada2014anomaly}, \cite{thompson2002implicit} or the comprehensive review of anomaly detection methods \cite{pimentel2014review}).

  It consists of two MLPs - an encoder and a decoder.  The encoder represents the mapping $e_{\phi}:\mathcal{X} \rightarrow \mathcal{Z}$  that projects a sample $x$ from the data space to code $z$ in the latent space. The decoder reconstructs the code back to the data space via mapping $d_{\theta}:\mathcal{Z} \rightarrow \mathcal{X}$. From here on, $\theta$ and $\phi$ denote hidden parameters of neural nets. Both parts of the autoencoder are trained using backpropagation by minimizing the reconstruction error 
    \begin{equation}
      \mathcal{L}_r(x,\phi,\theta) = || x - d_{\theta}(e_{\phi}(x)) ||^2_2.
    \end{equation}
  Because the dimension of the code $z$ is smaller than that of $x$, the autoencoder is forced to learn an efficient sparse representation of $x$ while being robust to noise \cite{vincent2010stacked}.

  This roughly equals to learning to reconstruct samples coming from the distribution $p(x)$ of the normal data. When the trained autoencoder is shown an anomaly, it will likely not produce a good reconstruction as it has not been trained with similar samples. Therefore, the anomaly score is given by the reconstruction error of a sample
  \begin{equation}
    f_{\text{AE}}(x) = \mathcal{L}_r(x,\bar{\phi},\bar{\theta}),
  \end{equation}
  where $\bar{\phi},\bar{\theta}$ are the learned parameters of the AE.

\subsubsection{Variational autoencoder}
  The variational autoencoder (VAE) \cite{kingma2013vae} and its modifications has seen a lot of success especially in generating realistic images. Its design is very similar to that of an  ordinary autoencoder. To induce the generative property, we force the encoder to produce codes $z$ that resemble samples from some easy--to--sample--from prior distribution $p(z)$ (e.g. $\mathcal{N}(0,1)$). Afterwards, new samples that resemble the real data can be generated by inputing codes sampled from $p(z)$ to the decoder.

  In VAE, both the encoder and decoder model parameters of conditional distributions denoted as $q_{\phi}(z|x)$ and $p_{\theta}(x|z)$. We assume that the distributions are Gaussian, therefore at the output layer of the MLPs we obtain the mean and variance of the respective distributions. During training, the aim is to minimize the reconstruction loss of $x$ while simultaneously minimizing the Kullback-Leibler divergence $D_{KL}(q_{\phi}(z|x)||p(z))$, which is zero if the two distributions are equal. Also, an important part of the training is the \textit{reparametrization trick}, which creates random decoder inputs in the following manner: $z=\mu_z + \sigma_z \epsilon$, where $\mu_z$ and $\sigma_z$ are outputs of the encoder and $\epsilon$ is sampled from $p(z)$. The cost function for training a VAE is following
    \begin{equation}
      \mathcal{L}_{v}(x,\phi,\theta) = \mathbb{E}_{q(z|x)} \left[ || x - d_{\theta}(z) ||^2_2 \right] + \lambda D_{KL}(q_{\phi}(z|x)||p(z)),
    \end{equation}
  where $\lambda = \sigma^2_x$ is a tuning parameter equal to a known variance of data. Since both $q_{\phi}(z|x)$ and $p(z)$ are Gaussian, there is an analytical expression for their KL divergence.
  
  There are numerous papers describing the use of VAE for anomaly detection --  \cite{solch2016variational}, \cite{xu2018unsupervised}, \cite{clachar2016novelty}, but none make a more complete comparison with other generative and classical methods and only some use non--image datasets. In \cite{an2015variational}, the authors describe the advantages of VAE over AE -- it generalizes more easily since it is working on probabilities. The anomaly score function of the VAE is the reconstruction error
  \begin{equation}
    f_{\text{VAE}}(x) = \mathbb{E}_{q(z|x)} \left[ || x - d_{\theta}(z) ||^2_2 \right].
  \end{equation}
  Alternatively, the log--likelihood of the code $z$ due to prior $p(z)$ can be used.

\subsection{Adversarial models}
\subsubsection{GAN}
The simplest adversarial generative model is the \textit{generative adversarial network} -- GAN \cite{goodfellow2014gan}. It consists of two adversaries -- a generator and a discriminator that are represented by MLPs. The generator creates samples that resemble the real data, while the discriminator is trying to recognize the fake samples from the real ones. During training, they both improve -- generator creates more believable samples while the discriminator gets more proficient at recognizing fakes. 

Inputs of the generator are codes $z \sim p(z)$, where $p(z)$ is e.g. standard or uniform distribution. The generator is a mapping into the data space $g_{\phi}:\mathcal{Z} \rightarrow \mathcal{X}$. Discriminator is a mapping $d_{\theta}:\mathcal{X} \rightarrow \left[0, 1\right]$, i.e. its output is a scalar representing the probability that a sample comes from the true data distribution $p(x)$. The training alternates between minimizing the logit cross--entropy for discriminator
  \begin{equation}
    \mathcal{L}_{d}(x,\theta) = - \mathbb{E}_{p(x)}\left[ \log d_{\theta}(x) \right] - \mathbb{E}_{p(z)}\left[ \log (1-d_{\theta}(g_{\phi}(z))) \right]
  \end{equation}
  and a simplified logit cross--entropy for the generator
  \begin{equation}
    \mathcal{L}_{g}(x,\phi) = - \mathbb{E}_{p(z)}\left[ \log d_{\theta}(g_{\phi}(z)) \right].
  \end{equation}
  After training, one can input samples $z$ to the generator and it should be able to generate samples that resemble those from $p(x)$.
  
  For anomaly detection application, $p(x)$ is the distribution of normal data. We do not need to know the true form of $p(x)$, we only need to be able to sample from it. The discriminator should ideally learn the however complicated form of $p(x)$ by backpropagation. Anomaly score of a sample $x$ is computed as a weighted average of the discriminator output and a reconstruction error of the generated sample
  \begin{equation}
    f_{\text{GAN}}(x) = -(1-\lambda)\log(d_{\theta}(x)) + \lambda || x - g_{\phi}(z) ||_2, 
  \end{equation}
  where $z\sim p(z)$ and $\lambda$ is a scalar scaling parameter.

\subsubsection{Feature-matching GAN}
  While GAN enjoys success in generation of realistic images, it is famously difficult to train. The authors of \cite{salimans2016fmgan} proposed a number of modifications to the original simple training process. One of them is the addition of the \textit{feature--matching} loss to the cost function. It is based on the idea that backpropagation based on a loss computed somewhere else than the final scalar output of the discriminator may provide improved gradients for the generator, thus enabling a more stable training procedure.
  
  The feature--matching GAN (fmGAN) uses an augmented generator cost function
  \begin{equation}
    \mathcal{L}_{f}(x,\phi) = \alpha \mathcal{L}_{g}(x,\phi) + \mathbb{E}_{p(x),p(z)}\left[ || h_{\theta}(x) - h_{\theta}(g_{\phi}(z)) ||_2 \right],
  \end{equation}
  where $h_{\theta}$ is the output of some intermediate (e.g. the penultimate) layer of the discriminator and $\alpha$ is a scalar scaling parameter. 
  
  The fmGAN has been successfully used for anomaly detection in \cite{schlegl2017unsupervised}. The anomaly score function is the same as in the case of the GAN model.

\subsection{Classical outlier detection methods}
\subsubsection{kNN}
  The $k$--nearest neighbours anomaly detection algorithm \cite{angiulli2002fast} is a simple yet powerful model. It is based on the assumption that normal data are grouped in the data space and anomalies are distant from them and therefore can be detected by measuring their distance from the rest of the data. It is relatively easy to implement, well described in literature and quite well performing. A large study \cite{campos2016evaluation} concluded that no classical anomaly detection algorithm provides a comprehensive improvement over kNN.
  
  For faster distance computation, the training data are encoded in a KDTree structure \cite{bentley1975multidimensional}. This significantly improves the prediction times on large datasets but requires additional overhead in construction of the tree. The only hyperparameter of the algorithm is $k$. The version of kNN used for experiments in this paper computes the anomaly score as the average distance to $k$--nearest neighbours in the training dataset. 

\subsubsection{Isolation forest}
 Isolation forest \cite{liu2008isolation} is also one of the more widely used anomaly detection algorithms. During training, the algorithm randomly and recursively partitions the data and stores this partitioning in a tree structure. This partitioning is done numerous times -- thus a forest is constructed. Anomalies should be ideally reached on a shorter path in the tree as they require a smaller number of partitionings to be separated from the rest of the data. The hyperparameter to be tuned is the number of trees $n_t$. The anomaly score is computed as the negative of the average path length of a sample over all trees in the forest.

\section{Experiments}
This section describes the setup of experiment that was conducted on a large number of benchmark datasets. Datasets, algorithm and experiment implementation in the Julia language is made publicly available in a GitHub repository at \verb+https://github.com/smidl/AnomalyDetection.jl+.

\subsection{Experiment setup}
\subsubsection{Benchmark data}
 In order to evaluate the selected models in the most realistic fashion, the methodology of creating appropriate anomaly detection benchmarks from \cite{emmott2013systematic} was adopted. A list of 35 preprocessed basic datasets from \cite{pevny2016loda} was used. A more detailed overview of the basic datasets and of the preprocessing can be found in the same paper.

 The basic datasets were originally created from multiclass classification or regression datasets that were split into normal and anomalous data, whitened and further processed. Afterwards, it is possible to sample normal and anomalous data according to different criteria. First, there are four levels of anomaly difficulty - \textit{easy, medium, hard} and \textit{very hard}. Second, a contamination rate (the percentage of anomalies in a sampled dataset) can be specified. Third, it is possible to sample anomalies that are clustered or unclustered.

 For the purposes of our experiments, we have sampled from each basic dataset to obtain a non-contaminated (containing no anomalies) training dataset and a contaminated testing dataset. We have used 80\% of normal data for training and 20\% for testing. In each of the basic datasets, there is a different number of anomalies of a given anomaly difficulty level (see the summary table in \cite{pevny2016loda}). Therefore, the anomaly difficulty levels were chosen (mostly \textit{easy} or \textit{medium}) so that there were enough anomalies to be sampled from with respect to the chosen contamination rate. The contamination rate of the testing dataset was 5\% and we sampled for non--clustered anomalies. Random sampling was repeated 10 times for each basic dataset to obtain diverse training and testing datasets .

\begin{table}
 \begin{tabular}[h]{c|c}
  algorithm & hyperparameters \\
  \hline
  kNN & $k \in \lbrace 1, \frac{1}{2}\sqrt{N}, \sqrt{N}, \frac{3}{2}\sqrt{N}, 2\sqrt{N} \rbrace$ \\
  
  Isolation Forest & $n_t \in \lbrace 100,  500 \rbrace$ \\
  
  all deep models &  $\text{dim}(\mathcal{Z}) \in \lbrace 0.1\tilde{M}, 0.2\tilde{M}, \dots, 0.5\tilde{M} \rbrace$, \\ 
     & $\tilde{M} = \min(200, M)$ \\
    & $n_h \in \lbrace 1,2,3 \rbrace$ \\
  VAE & $\lambda \in \lbrace 10^{-4}, 10^{-3}, \dots, 1 \rbrace$ \\
  GAN & $\lambda \in \lbrace 0, 0.2, \dots, 1.0 \rbrace$ \\
  fmGAN & $\alpha \in \lbrace 10^{-4}, 10^{-2}, \dots, 10^4 \rbrace$ \\ 
   & $\lambda \in \lbrace 0, 0.2, \dots, 1.0 \rbrace$ 
 \end{tabular}
 \caption{Tested hyperparameter settings. $M$ is the dimensionality of a dataset, $N$ is the number of training samples, $n_t$ is the number of trees, $n_h$ is the number of hidden layers.} 
 \label{tab:hyperparams}
\end{table}

 \subsubsection{Algorithm setup}
 The table \ref{tab:hyperparams} contains an overview of tested hyperparameter values. An experiment consisting of training of a model and predicting anomaly scores was carried out for every possible combination of hyperparameters. Therefore, a different number of experiments per dataset sampling was realized for different algorithms -- that is no more than 5 experiments for kNN but up to 450 experiments in the case of fmGAN. For training of the deep models, the Adam optimizer \cite{kingma2014adam} was used with a learning rate of 0.001. We trained each generative model for a total of 10000 steps with a batchsize of 256 or lower in case of datasets that are not numerous enough. Relu activation functions and dense layers were used. Size of hidden layers was given by linear interpolation between the code dimension $\text{dim}(\mathcal{Z})$ and a proper input/output dimension. The code dimension was proportionate to the input dimension $\text{dim}(\mathcal{X})$ but never exceeded 100.

\subsection{Evaluation methodology}
\subsubsection{Algorithm ranking}
For comparison of different algorithms on multiple datasets, we use the statistical procedure described in \cite{demsar2006statistical}. For a given comparison metric, e.g. AUROC (area under ROC curve), we rank algorithms according to their performance on a dataset. If there is a tie, then average rank is given to the tied algorithms. Afterwards, an average rank over datasets is computed for each algorithm. See e.g. Table \ref{tab:testaucfull}, where the evaluation metric is the maximum AUROC on the testing dataset (the first ranking criterion in the following text).  Based on the average ranks, a number of statistical hypotheses may be tested. In the case of this study, we are interested in two: a) is there enough statistical evidence to reject the hypothesis that all the algorithms perform equally and b) do two individual algorithms perform differently on a statistically significant level? The answers to these questions are given by the Friedman and Nemenyi test, respectively. For our problem size (6 algorithms and 35 datasets), the critical value for the Nemenyi test at 5\% confidence level is $CD_{0.05}\approx1.2745$. If the average rank of two algorithms is larger than that, we can say that their performance is significantly different.

\begin{figure}[h] 
 \center 
 \begin{tikzpicture} 
  \draw (1.0,0) -- (6.0,0); 
  \foreach \x in {1,...,6} \draw (\x,0.10) -- (\x,-0.10) node[anchor=north]{$\x$}; 
  \draw (1.985714,0) -- (1.985714,0.19999999999999998) -- (0.9, 0.19999999999999998) node[anchor=east] {fmGAN}; 
  \draw (2.071429,0) -- (2.071429,0.5) -- (0.9, 0.5) node[anchor=east] {VAE}; 
  \draw (3.471429,0) -- (3.471429,0.7999999999999999) -- (0.9, 0.7999999999999999) node[anchor=east] {AE}; 
  \draw (3.9,0) -- (3.9,0.8) -- (6.1, 0.8) node[anchor=west] {GAN}; 
  \draw (3.942857,0) -- (3.942857,0.5) -- (6.1, 0.5) node[anchor=west] {kNN}; 
  \draw (5.628571,0) -- (5.628571,0.2) -- (6.1, 0.2) node[anchor=west] {IForest}; 
  \draw[line width=0.06cm,color=black,draw opacity=1.0] (1.955714,0.05) -- (2.101429,0.05); 
  \draw[line width=0.06cm,color=black,draw opacity=1.0] (3.4414290000000003,0.05) -- (3.972857,0.05); 
 \end{tikzpicture} 
 \caption{Critical difference diagram for the first hyperparameter selection criterion.} 
 \label{fig:cdd1} 
\end{figure}
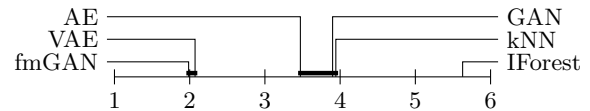

A concise way of visualization are the critical difference diagrams, see e.g. Fig \ref{fig:cdd1}. The average algorithm ranks are marked and the statistically equally performing algorithms are connected together with a wide black line.


\subsubsection{Hyperparameter selection}
After training of an algorithm on the training dataset, a vector of anomaly scores was computed for the testing dataset. Also, a small sample of anomalies was added to the training dataset and a vector of anomaly scores of the resulting ensemble was computed (reasoning behind this follows). Afterwards, testing and training AUROC were computed from these two vectors and known labels. This was done for each individual experiment specified by a combination of dataset, model, resampling iteration and model hyperparameter setting.

We assumed that tuning of hyperparameters is done during training. The rankings of models were produced using the testing AUROC of best--performant model in a resampling iteration, averaged over resampling iterations on a dataset.  To simulate different real--world conditions in which hyperparameters are tuned, 3 methods of their selection were used.
\begin{enumerate}
  \item Firstly, the hyperparameters were selected by the best AUROC on the testing dataset. This gives an idea of the maximum potential of the algorithm, which may be difficult to attain in reality. See Table \ref{tab:testaucfull}.
  \item Secondly, the hyperparameters were selected by best performance on training dataset with added anomalies (training AUROC). This simulates a more realistic scenario of hyperparameter tuning, where all the labels are known on the training dataset. See Table \ref{tab:trainaucfull}.
  \item  Thirdly, the hyperparameters were selected using the $p$\% most anomalous samples of the training dataset with added anomalies. However, the performance was measured using the precision (rate of detected anomalies). This is the most realistic scenario in case when the training labels are expensive to obtain. The common practice is then to manually inspect the most $p$\% anomalous samples in the training dataset and select the hyperparameters of the most precise model. For our experiments, we chose $p\in\lbrace 1, 5 \rbrace$. See Table \ref{tab:top5aucfull} and \ref{tab:top1aucfull}.
\end{enumerate}

These ranking methods stimulate interesting questions.
\begin{itemize}
  \item What is the reason of an algorithm performing well in the first method and not so well in the others? Insufficient tuning of hyperparameters? Overfitting on training? Or just very difficult applicability to real--world problems where labels are not present and such tuning is not possible?
  \item What is the robustness of different algorithms to variation of the training data? How robust is it to a possible presence of anomalies in the training data?
  \item How many samples need to be manually inspected to obtain robust optimal hyperparameter settings?
  \item Can we analyze on which datasets are some methods more successful than the others? Do these datasets have something in common?
\end{itemize}

\begin{table} 
 \center 
 \begin{tabular}[h]{c | c c c c c c } 
    & kNN & IForest & AE & VAE & GAN & fmGAN  \\ 
  \hline 
  test auc & 3.94 & 5.63 & 3.47 & 2.07 & 3.90 & 1.99  \\ 
  train auc & 3.13 & 4.61 & 3.63 & 2.84 & 4.46 & 2.33  \\ 
  top 5\% & 2.57 & 4.07 & 3.24 & 2.73 & 4.90 & 3.49  \\ 
  top 1\% & 2.14 & 3.53 & 3.13 & 2.93 & 4.97 & 4.30  \\ 
 \end{tabular}
 \caption{Average ranks of algorithms for different hyperparameter selection criteria.} 
 \label{tab:aucsummary} 
\end{table}
\begin{table} 
 \center 
 \begin{tabular}[h]{c | c c c c c c } 
    & kNN & IForest & AE & VAE & GAN & fmGAN  \\ 
  \hline 
  $t_f$ [s] & 0.15 & 0.76 & 276.33 & 1088.71 & 844.83 & 879.94  \\ 
  $t_p$ [s] & 5.35 & 0.58 & 2.14 & 3.49 & 6.40 & 6.38  \\ 
 \end{tabular}
 \caption{Average fit $t_f$ and predict $t_p$ times.} 
 \label{tab:timesummary} 
\end{table}

\subsection{Experimental results}
Mean ranks of algorithms across datasets for all hyperparameter selection methods are summarized in Table \ref{tab:aucsummary} (the complete AUC scores are shown in Tables \ref{tab:testaucfull} -- \ref{tab:top1aucfull} in Appendix). Results from Table~\ref{tab:aucsummary} shows that generative models, namely VAE and fmGAN, can produce excellent results if one has a large number of known anomalies to select right hyperparameters (methods \textit{test auc} and \textit{train auc}). Once the number of known anomalies decreases, as is the case of the more realistic methods \textit{top 1\%} and \textit{top 5\%}, generative models becomes inferior to methods robust to hyperparameter selection, namely k--nearest neighbours.

The Friedman test rejects the hypothesis that the models perform equally well in all four ranking criteria. Critical difference diagrams shown in Figs. \ref{fig:cdd1} -- \ref{fig:cdd4} reveal that unless one selects hyperparameters on testing set, which is obviously cheating, not a single generative model provides a statistically significant improvement over the naive kNN anomaly detection algorithm. VAE seems to be the most robust and therefore more promising deep model, as its ranks relatively well for all the hyperparameter selection criteria. 

Although the deep models do not generally outperform kNN when tuned with a limited number of labels, they can still perform well on certain datasets. See Table \ref{tab:top1aucfull}, datasets \textit{cardiotocography}, \textit{libras} and \textit{sonar} for the ranks of fmGAN. In Table \ref{tab:top5aucfull}, fmGAN against performs well on these (and additional) datasets. The same can be observed for VAE and the \textit{ecoli, miniboone} and \textit{multiple--features} datasets. This may mean that these datasets share some common characteristic which makes them suitable for application of deep models, rather than just a random fluctuation caused by the large number of experiments carried out for fmGAN and VAE. However we have not yet been able to find the connection. It is certainly not the difficulty of the anomalies, as kNN performs relatively well across all difficulty levels.

The only area in which the deep models outperform kNN is the mean prediction time on large datasets (in terms of number of training samples), see Table \ref{tab:timesummary}. Note that the largest benchmark dataset is \textit{miniboone} with 93565 normal samples. In deep models, the prediction is independent on the training data size. This is however compensated for by the computational demands of their training.

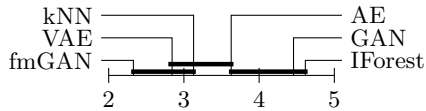
\begin{figure}[h] 
 \center 
 \begin{tikzpicture} 
  \draw (2.0,0) -- (5.0,0); 
  \foreach \x in {2,...,5} \draw (\x,0.10) -- (\x,-0.10) node[anchor=north]{$\x$}; 
  \draw (2.328571,0) -- (2.328571,0.19999999999999998) -- (1.9, 0.19999999999999998) node[anchor=east] {fmGAN}; 
  \draw (2.842857,0) -- (2.842857,0.5) -- (1.9, 0.5) node[anchor=east] {VAE}; 
  \draw (3.128571,0) -- (3.128571,0.7999999999999999) -- (1.9, 0.7999999999999999) node[anchor=east] {kNN}; 
  \draw (3.628571,0) -- (3.628571,0.8) -- (5.1, 0.8) node[anchor=west] {AE}; 
  \draw (4.457143,0) -- (4.457143,0.5) -- (5.1, 0.5) node[anchor=west] {GAN}; 
  \draw (4.614286,0) -- (4.614286,0.2) -- (5.1, 0.2) node[anchor=west] {IForest}; 
  \draw[line width=0.06cm,color=black,draw opacity=1.0] (2.2985710000000004,0.05) -- (3.158571,0.05); 
  \draw[line width=0.06cm,color=black,draw opacity=1.0] (2.842857-0.05,0.15000000000000002) -- (3.658571,0.15000000000000002); 
  \draw[line width=0.06cm,color=black,draw opacity=1.0] (3.598571,0.05) -- (4.644286,0.05); 
 \end{tikzpicture} 
 \caption{Critical difference diagram for the second hyperparameter selection criterion.} 
 \label{fig:cdd2} 
\end{figure}
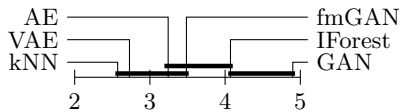
\begin{figure}[h] 
 \center 
 \begin{tikzpicture} 
  \draw (2.0,0) -- (5.0,0); 
  \foreach \x in {2,...,5} \draw (\x,0.10) -- (\x,-0.10) node[anchor=north]{$\x$}; 
  \draw (2.571429,0) -- (2.571429,0.19999999999999998) -- (1.9, 0.19999999999999998) node[anchor=east] {kNN}; 
  \draw (2.728571,0) -- (2.728571,0.5) -- (1.9, 0.5) node[anchor=east] {VAE}; 
  \draw (3.242857,0) -- (3.242857,0.7999999999999999) -- (1.9, 0.7999999999999999) node[anchor=east] {AE}; 
  \draw (3.485714,0) -- (3.485714,0.8) -- (5.1, 0.8) node[anchor=west] {fmGAN}; 
  \draw (4.071429,0) -- (4.071429,0.5) -- (5.1, 0.5) node[anchor=west] {IForest}; 
  \draw (4.9,0) -- (4.9,0.2) -- (5.1, 0.2) node[anchor=west] {GAN}; 
  \draw[line width=0.06cm,color=black,draw opacity=1.0] (2.5414290000000004,0.05) -- (3.515714,0.05); 
  \draw[line width=0.06cm,color=black,draw opacity=1.0] (3.242857-0.05,0.15000000000000002) -- (4.101429,0.15000000000000002); 
  \draw[line width=0.06cm,color=black,draw opacity=1.0] (4.041429,0.05) -- (4.930000000000001,0.05); 
 \end{tikzpicture} 
 \caption{Critical difference diagram for the third hyperparameter selection criterion at 5\% most anomalous samples.} 
 \label{fig:cdd3} 
\end{figure}
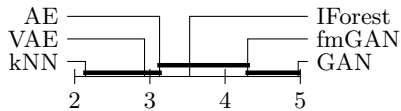
\begin{figure}[h] 
 \center 
 \begin{tikzpicture} 
  \draw (2.0,0) -- (5.0,0); 
  \foreach \x in {2,...,5} \draw (\x,0.10) -- (\x,-0.10) node[anchor=north]{$\x$}; 
  \draw (2.142857,0) -- (2.142857,0.19999999999999998) -- (1.9, 0.19999999999999998) node[anchor=east] {kNN}; 
  \draw (2.928571,0) -- (2.928571,0.5) -- (1.9, 0.5) node[anchor=east] {VAE}; 
  \draw (3.128571,0) -- (3.128571,0.7999999999999999) -- (1.9, 0.7999999999999999) node[anchor=east] {AE}; 
  \draw (3.528571,0) -- (3.528571,0.8) -- (5.1, 0.8) node[anchor=west] {IForest}; 
  \draw (4.3,0) -- (4.3,0.5) -- (5.1, 0.5) node[anchor=west] {fmGAN}; 
  \draw (4.971429,0) -- (4.971429,0.2) -- (5.1, 0.2) node[anchor=west] {GAN}; 
  \draw[line width=0.06cm,color=black,draw opacity=1.0] (2.112857,0.05) -- (3.158571,0.05); 
  \draw[line width=0.06cm,color=black,draw opacity=1.0] (3.098571,0.15000000000000002) -- (4.33,0.15000000000000002); 
  \draw[line width=0.06cm,color=black,draw opacity=1.0] (4.27,0.05) -- (5.001429,0.05); 
 \end{tikzpicture} 
 \caption{Critical difference diagram for the third hyperparameter selection criterion at 1\% most anomalous samples.} 
 \label{fig:cdd4} 
\end{figure}

\section{Conclusions}
In this paper, a selection of deep generative models adopted for anomaly detection were compared against traditional methods of k-nearest neighbours and Isolation Forests. This was done in a systematic way on a large number of benchmark datasets. Most authors of novel anomaly detection methods based on deep models do not compare to kNN, but instead compare to methods such as Local Outlier Factor \cite{breunig2000lof}, despite it has been shown in \cite{pevny2016loda} and \cite{campos2016evaluation} that it does not systematically outperform kNN algorithm. We have shown that the robustness of kNN still holds even in comparison with some deep generative models.

From the conducted experiments, the main conclusion is that the major bottleneck for reliable employment of deep generative models for anomaly detection is the difficulty of hyperparameter tuning. Ideally, hyperparameter tuning should be automated and included in the training procedure, which is not the case. Especially adversarial models do not seem to be robust and their training is difficult. On the other hand, generative models showed promising potential and should be studied in a greater depth. We have studied performance of different models under a relatively simple, yet realistic, hyperparameter selection methods and criteria. An interesting question is the existence of a criterion that is more appropriate in the context of limited number of available labels.

This also leads us to questioning the appropriateness of the use of deep generative models in the unsupervised setting. Since we have shown that labels are necessary for hyperparameter tuning, maybe it is worth investigating a semi/fully--supervised deep generative models for anomaly detection.

\section{Acknowledgments}
Research presented in this work has been supported by the Grant SGS18/188/OHK4/3T/14 provided by the Ministry of Education, Youth, and Sports of the Czech Republic (M\v SMT \v CR) and by the GACR project 18-21409S.

%
\bibliographystyle{abbrv}
\bibliography{bib}  

\begin{thebibliography}{10}

\bibitem{an2015variational}
J.~An and S.~Cho.
\newblock Variational autoencoder based anomaly detection using reconstruction
  probability.
\newblock {\em SNU Data Mining Center, Tech. Rep.}, 2015.

\bibitem{angiulli2002fast}
F.~Angiulli and C.~Pizzuti.
\newblock Fast outlier detection in high dimensional spaces.
\newblock In {\em European Conference on Principles of Data Mining and
  Knowledge Discovery}, pages 15--27. Springer, 2002.

\bibitem{bentley1975multidimensional}
J.~L. Bentley.
\newblock Multidimensional binary search trees used for associative searching.
\newblock {\em Communications of the ACM}, 18(9):509--517, 1975.

\bibitem{breunig2000lof}
M.~M. Breunig, H.-P. Kriegel, R.~T. Ng, and J.~Sander.
\newblock Lof: identifying density-based local outliers.
\newblock In {\em ACM sigmod record}, volume~29, pages 93--104. ACM, 2000.

\bibitem{campos2016evaluation}
G.~O. Campos, A.~Zimek, J.~Sander, R.~J. Campello, B.~Micenkov{\'a},
  E.~Schubert, I.~Assent, and M.~E. Houle.
\newblock On the evaluation of unsupervised outlier detection: measures,
  datasets, and an empirical study.
\newblock {\em Data Mining and Knowledge Discovery}, 30(4):891--927, 2016.

\bibitem{clachar2016novelty}
S.~Clachar.
\newblock {\em Novelty detection and cluster analysis in time series data using
  variational autoencoder feature maps}.
\newblock The University of North Dakota, 2016.

\bibitem{demsar2006statistical}
J.~Dem{\v{s}}ar.
\newblock Statistical comparisons of classifiers over multiple data sets.
\newblock {\em Journal of Machine learning research}, 7(Jan):1--30, 2006.

\bibitem{emmott2013systematic}
A.~F. Emmott, S.~Das, T.~Dietterich, A.~Fern, and W.-K. Wong.
\newblock Systematic construction of anomaly detection benchmarks from real
  data.
\newblock In {\em Proceedings of the ACM SIGKDD workshop on outlier detection
  and description}, pages 16--21. ACM, 2013.

\bibitem{goodfellow2014gan}
I.~Goodfellow, J.~Pouget-Abadie, M.~Mirza, B.~Xu, D.~Warde-Farley, S.~Ozair,
  A.~Courville, and Y.~Bengio.
\newblock Generative adversarial nets.
\newblock In {\em Advances in neural information processing systems}, pages
  2672--2680, 2014.

\bibitem{kingma2014adam}
D.~P. Kingma and J.~Ba.
\newblock Adam: A method for stochastic optimization.
\newblock {\em arXiv preprint arXiv:1412.6980}, 2014.

\bibitem{kingma2013vae}
D.~P. Kingma and M.~Welling.
\newblock Auto-encoding variational bayes.
\newblock {\em arXiv preprint arXiv:1312.6114}, 2013.

\bibitem{liu2008isolation}
F.~T. Liu, K.~M. Ting, and Z.-H. Zhou.
\newblock Isolation forest.
\newblock In {\em Data Mining, 2008. ICDM'08. Eighth IEEE International
  Conference on}, pages 413--422. IEEE, 2008.

\bibitem{odena2018realistic}
A.~Odena, A.~Oliver, C.~Raffel, E.~D. Cubuk, and I.~Goodfellow.
\newblock Realistic evaluation of semi-supervised learning algorithms.
\newblock 2018.

\bibitem{pevny2016loda}
T.~Pevn{\'y}.
\newblock Loda: Lightweight on-line detector of anomalies.
\newblock {\em Machine Learning}, 102(2):275--304, 2016.

\bibitem{pimentel2014review}
M.~A. Pimentel, D.~A. Clifton, L.~Clifton, and L.~Tarassenko.
\newblock A review of novelty detection.
\newblock {\em Signal Processing}, 99:215--249, 2014.

\bibitem{sakurada2014anomaly}
M.~Sakurada and T.~Yairi.
\newblock Anomaly detection using autoencoders with nonlinear dimensionality
  reduction.
\newblock In {\em Proceedings of the MLSDA 2014 2nd Workshop on Machine
  Learning for Sensory Data Analysis}, page~4. ACM, 2014.

\bibitem{salimans2016fmgan}
T.~Salimans, I.~Goodfellow, W.~Zaremba, V.~Cheung, A.~Radford, and X.~Chen.
\newblock Improved techniques for training gans.
\newblock In {\em Advances in Neural Information Processing Systems}, pages
  2234--2242, 2016.

\bibitem{schlegl2017unsupervised}
T.~Schlegl, P.~Seeb{\"o}ck, S.~M. Waldstein, U.~Schmidt-Erfurth, and G.~Langs.
\newblock Unsupervised anomaly detection with generative adversarial networks
  to guide marker discovery.
\newblock In {\em International Conference on Information Processing in Medical
  Imaging}, pages 146--157. Springer, 2017.

\bibitem{solch2016variational}
M.~S{\"o}lch, J.~Bayer, M.~Ludersdorfer, and P.~van~der Smagt.
\newblock Variational inference for on-line anomaly detection in
  high-dimensional time series.
\newblock {\em arXiv preprint arXiv:1602.07109}, 2016.

\bibitem{thompson2002implicit}
B.~B. Thompson, R.~J. Marks, J.~J. Choi, M.~A. El-Sharkawi, M.-Y. Huang, and
  C.~Bunje.
\newblock Implicit learning in autoencoder novelty assessment.
\newblock In {\em Neural Networks, 2002. IJCNN'02. Proceedings of the 2002
  International Joint Conference on}, volume~3, pages 2878--2883. IEEE, 2002.

\bibitem{vincent2010stacked}
P.~Vincent, H.~Larochelle, I.~Lajoie, Y.~Bengio, and P.-A. Manzagol.
\newblock Stacked denoising autoencoders: Learning useful representations in a
  deep network with a local denoising criterion.
\newblock {\em Journal of Machine Learning Research}, 11(Dec):3371--3408, 2010.

\bibitem{xu2018unsupervised}
H.~Xu, W.~Chen, N.~Zhao, Z.~Li, J.~Bu, Z.~Li, Y.~Liu, Y.~Zhao, D.~Pei, Y.~Feng,
  et~al.
\newblock Unsupervised anomaly detection via variational auto-encoder for
  seasonal kpis in web applications.
\newblock {\em arXiv preprint arXiv:1802.03903}, 2018.

\bibitem{zhai2016deep}
S.~Zhai, Y.~Cheng, W.~Lu, and Z.~Zhang.
\newblock Deep structured energy based models for anomaly detection.
\newblock {\em arXiv preprint arXiv:1605.07717}, 2016.

\bibitem{zong2018deep}
B.~Zong, Q.~Song, M.~R. Min, W.~Cheng, C.~Lumezanu, D.~Cho, and H.~Chen.
\newblock Deep autoencoding gaussian mixture model for unsupervised anomaly
  detection.
\newblock In {\em International Conference on Learning Representations}, 2018.

\end{thebibliography}
%
%
\appendix
Here, selected results of scores and ranks for algorithms across all datasets are shown. For brevity, we use short codes for dataset names, see Table \ref{tab:codes}.

\begin{table} 
 \center 
 \resizebox{\columnwidth}{!}{ 
 \begin{tabular}[h]{c c | c c } 
  abalone & aba & page-blocks & pag  \\ 
  blood-transfusion & blo & parkinsons & par  \\ 
  breast-cancer-wisconsin & brc & pendigits & pen  \\ 
  breast-tissue & brt & pima-indians & pim  \\ 
  cardiotocography & car & sonar & son  \\ 
  ecoli & eco & spect-heart & spe  \\ 
  glass & gla & statlog-satimage & ssa  \\ 
  haberman & hab & statlog-segment & sse  \\ 
  ionosphere & ion & statlog-shuttle & ssh  \\ 
  iris & iri & statlog-vehicle & sve  \\ 
  isolet & iso & synthetic-control-chart & syn  \\ 
  letter-recognition & let & vertebral-column & ver  \\ 
  libras & lib & wall-following-robot & wal  \\ 
  madelon & mad & waveform-1 & wa1  \\ 
  magic-telescope & mag & waveform-2 & wa2  \\ 
  miniboone & min & wine & win  \\ 
  multiple-features & mul & yeast & yea  \\ 
  musk-2 & mus &  &   \\ 
 \end{tabular}
 }
 \caption{Overview of coded dataset names used in Tables \ref{tab:testaucfull} -- \ref{tab:top1aucfull}.} 
 \label{tab:codes} 
\end{table}

\newpage
\newpage
\begin{table} 
 \center 
 \resizebox{\columnwidth}{!}{ 
 \begin{tabular}[h]{c | c c c c c c } 
  dataset & kNN & IForest & AE & VAE & GAN & fmGAN  \\ 
  \hline 
  aba & 0.93(3.5) & 0.83(6.0) & 0.94(1.5) & 0.94(1.5) & 0.88(5.0) & 0.93(3.5)  \\ 
  blo & 0.99(4.0) & 0.98(6.0) & 0.99(4.0) & 1.00(1.5) & 0.99(4.0) & 1.00(1.5)  \\ 
  brc & 0.94(6.0) & 0.98(4.5) & 0.99(2.0) & 0.99(2.0) & 0.98(4.5) & 0.99(2.0)  \\ 
  brt & 0.99(5.5) & 0.99(5.5) & 1.00(2.5) & 1.00(2.5) & 1.00(2.5) & 1.00(2.5)  \\ 
  car & 0.66(6.0) & 0.70(4.0) & 0.67(5.0) & 0.83(3.0) & 0.85(2.0) & 0.91(1.0)  \\ 
  eco & 0.95(5.0) & 0.88(6.0) & 0.97(4.0) & 0.99(2.0) & 0.98(3.0) & 1.00(1.0)  \\ 
  gla & 0.90(5.0) & 0.86(6.0) & 0.98(4.0) & 1.00(2.0) & 1.00(2.0) & 1.00(2.0)  \\ 
  hab & 0.98(3.0) & 0.95(6.0) & 0.97(4.5) & 0.99(2.0) & 0.97(4.5) & 1.00(1.0)  \\ 
  ion & 0.96(5.0) & 0.94(6.0) & 0.99(3.5) & 1.00(1.5) & 0.99(3.5) & 1.00(1.5)  \\ 
  iri & 0.98(5.0) & 0.94(6.0) & 1.00(2.5) & 1.00(2.5) & 1.00(2.5) & 1.00(2.5)  \\ 
  iso & 0.78(3.0) & 0.59(6.0) & 0.76(4.0) & 0.81(2.0) & 0.64(5.0) & 0.83(1.0)  \\ 
  let & 0.80(3.0) & 0.57(6.0) & 0.79(4.0) & 0.83(1.0) & 0.62(5.0) & 0.81(2.0)  \\ 
  lib & 0.88(5.0) & 0.75(6.0) & 0.91(4.0) & 0.98(2.0) & 0.96(3.0) & 0.99(1.0)  \\ 
  mad & 0.59(6.0) & 0.60(5.0) & 0.65(4.0) & 0.71(2.5) & 0.71(2.5) & 0.79(1.0)  \\ 
  mag & 0.95(1.0) & 0.86(5.5) & 0.93(2.5) & 0.93(2.5) & 0.86(5.5) & 0.88(4.0)  \\ 
  min & 0.89(4.0) & 0.85(5.0) & 0.76(6.0) & 0.93(1.5) & 0.91(3.0) & 0.93(1.5)  \\ 
  mul & 0.99(2.0) & 0.81(6.0) & 0.99(2.0) & 0.99(2.0) & 0.85(5.0) & 0.97(4.0)  \\ 
  mus & 0.97(3.0) & 0.61(6.0) & 0.99(1.5) & 0.99(1.5) & 0.89(5.0) & 0.93(4.0)  \\ 
  pag & 0.98(3.5) & 0.96(5.0) & 0.99(1.5) & 0.99(1.5) & 0.95(6.0) & 0.98(3.5)  \\ 
  par & 0.94(5.0) & 0.87(6.0) & 0.98(4.0) & 0.99(2.5) & 0.99(2.5) & 1.00(1.0)  \\ 
  pen & 1.00(1.0) & 0.91(4.5) & 0.99(2.5) & 0.99(2.5) & 0.86(6.0) & 0.91(4.5)  \\ 
  pim & 0.91(4.5) & 0.88(6.0) & 0.92(3.0) & 0.94(2.0) & 0.91(4.5) & 0.98(1.0)  \\ 
  son & 0.84(5.0) & 0.81(6.0) & 0.95(4.0) & 1.00(2.0) & 1.00(2.0) & 1.00(2.0)  \\ 
  spe & 0.91(5.0) & 0.88(6.0) & 0.97(4.0) & 1.00(2.0) & 1.00(2.0) & 1.00(2.0)  \\ 
  ssa & 0.97(1.0) & 0.92(5.5) & 0.96(3.0) & 0.96(3.0) & 0.92(5.5) & 0.96(3.0)  \\ 
  sse & 0.96(3.0) & 0.85(6.0) & 0.95(4.0) & 0.97(1.5) & 0.92(5.0) & 0.97(1.5)  \\ 
  ssh & 1.00(2.0) & 0.96(5.0) & 1.00(2.0) & 1.00(2.0) & 0.94(6.0) & 0.98(4.0)  \\ 
  sve & 0.84(5.0) & 0.79(6.0) & 0.91(3.0) & 0.93(2.0) & 0.90(4.0) & 0.95(1.0)  \\ 
  syn & 0.98(3.5) & 0.89(6.0) & 0.98(3.5) & 1.00(1.5) & 0.96(5.0) & 1.00(1.5)  \\ 
  ver & 0.82(4.5) & 0.79(6.0) & 0.82(4.5) & 0.92(2.0) & 0.90(3.0) & 0.98(1.0)  \\ 
  wal & 0.86(2.0) & 0.73(5.0) & 0.79(4.0) & 0.80(3.0) & 0.72(6.0) & 0.87(1.0)  \\ 
  wa1 & 0.82(4.0) & 0.81(5.0) & 0.80(6.0) & 0.86(2.0) & 0.85(3.0) & 0.91(1.0)  \\ 
  wa2 & 0.82(4.0) & 0.80(5.5) & 0.80(5.5) & 0.86(3.0) & 0.87(2.0) & 0.94(1.0)  \\ 
  win & 0.98(5.0) & 0.96(6.0) & 1.00(2.5) & 1.00(2.5) & 1.00(2.5) & 1.00(2.5)  \\ 
  yea & 0.84(5.0) & 0.80(6.0) & 0.91(3.0) & 0.93(2.0) & 0.86(4.0) & 0.95(1.0)  \\ 
  \hline
  avg & 0.90(3.94) & 0.84(5.63) & 0.91(3.47) & 0.94(2.07) & 0.90(3.9) & 0.95(1.99)  \\ 
 \end{tabular}
 }
 \caption{AUROC scores and ranks of algorithms using the first hyperparameter selection criterion. The last line is an average.} 
 \label{tab:testaucfull} 
\end{table}
\begin{table} 
 \center 
 \resizebox{\columnwidth}{!}{ 
 \begin{tabular}[h]{c | c c c c c c } 
  dataset & kNN & IForest & AE & VAE & GAN & fmGAN  \\ 
  \hline 
  aba & 0.93(1.0) & 0.83(6.0) & 0.92(2.5) & 0.92(2.5) & 0.84(5.0) & 0.86(4.0)  \\ 
  blo & 0.98(4.0) & 0.98(4.0) & 0.98(4.0) & 0.98(4.0) & 0.98(4.0) & 1.00(1.0)  \\ 
  brc & 0.93(6.0) & 0.98(1.0) & 0.95(3.0) & 0.94(4.5) & 0.94(4.5) & 0.96(2.0)  \\ 
  brt & 0.99(2.0) & 0.99(2.0) & 0.99(2.0) & 0.98(4.0) & 0.95(5.5) & 0.95(5.5)  \\ 
  car & 0.66(5.0) & 0.70(4.0) & 0.62(6.0) & 0.80(3.0) & 0.85(2.0) & 0.89(1.0)  \\ 
  eco & 0.93(3.0) & 0.87(5.5) & 0.94(1.5) & 0.94(1.5) & 0.87(5.5) & 0.89(4.0)  \\ 
  gla & 0.90(5.0) & 0.85(6.0) & 0.91(4.0) & 0.92(3.0) & 0.96(2.0) & 0.98(1.0)  \\ 
  hab & 0.97(1.0) & 0.94(3.5) & 0.86(5.0) & 0.94(3.5) & 0.83(6.0) & 0.95(2.0)  \\ 
  ion & 0.94(4.5) & 0.94(4.5) & 0.95(3.0) & 0.97(2.0) & 0.90(6.0) & 0.98(1.0)  \\ 
  iri & 0.97(5.0) & 0.92(6.0) & 0.98(4.0) & 0.99(2.5) & 0.99(2.5) & 1.00(1.0)  \\ 
  iso & 0.78(2.0) & 0.59(5.0) & 0.71(4.0) & 0.72(3.0) & 0.58(6.0) & 0.82(1.0)  \\ 
  let & 0.80(1.5) & 0.57(5.0) & 0.77(3.0) & 0.76(4.0) & 0.55(6.0) & 0.80(1.5)  \\ 
  lib & 0.87(2.5) & 0.75(6.0) & 0.82(5.0) & 0.83(4.0) & 0.92(1.0) & 0.87(2.5)  \\ 
  mad & 0.57(5.0) & 0.59(3.0) & 0.56(6.0) & 0.58(4.0) & 0.61(2.0) & 0.62(1.0)  \\ 
  mag & 0.95(1.0) & 0.86(5.0) & 0.93(2.5) & 0.93(2.5) & 0.85(6.0) & 0.88(4.0)  \\ 
  min & 0.75(6.0) & 0.85(4.0) & 0.76(5.0) & 0.93(1.5) & 0.91(3.0) & 0.93(1.5)  \\ 
  mul & 0.99(1.5) & 0.80(5.0) & 0.98(3.0) & 0.99(1.5) & 0.71(6.0) & 0.93(4.0)  \\ 
  mus & 0.97(3.0) & 0.61(6.0) & 0.98(2.0) & 0.99(1.0) & 0.89(5.0) & 0.93(4.0)  \\ 
  pag & 0.89(6.0) & 0.96(4.0) & 0.98(2.5) & 0.99(1.0) & 0.94(5.0) & 0.98(2.5)  \\ 
  par & 0.94(1.0) & 0.85(5.0) & 0.91(2.0) & 0.83(6.0) & 0.90(3.0) & 0.87(4.0)  \\ 
  pen & 1.00(1.0) & 0.91(4.0) & 0.98(3.0) & 0.99(2.0) & 0.86(6.0) & 0.88(5.0)  \\ 
  pim & 0.89(2.0) & 0.87(4.0) & 0.86(5.5) & 0.88(3.0) & 0.86(5.5) & 0.92(1.0)  \\ 
  son & 0.81(4.5) & 0.81(4.5) & 0.80(6.0) & 0.83(3.0) & 0.96(1.0) & 0.93(2.0)  \\ 
  spe & 0.89(1.5) & 0.86(5.0) & 0.89(1.5) & 0.88(3.0) & 0.85(6.0) & 0.87(4.0)  \\ 
  ssa & 0.97(1.0) & 0.92(5.0) & 0.95(3.5) & 0.95(3.5) & 0.90(6.0) & 0.96(2.0)  \\ 
  sse & 0.96(1.5) & 0.85(5.5) & 0.94(4.0) & 0.96(1.5) & 0.85(5.5) & 0.95(3.0)  \\ 
  ssh & 1.00(2.0) & 0.96(5.0) & 1.00(2.0) & 1.00(2.0) & 0.87(6.0) & 0.97(4.0)  \\ 
  sve & 0.83(4.0) & 0.78(6.0) & 0.84(3.0) & 0.85(2.0) & 0.82(5.0) & 0.93(1.0)  \\ 
  syn & 0.97(1.5) & 0.89(5.0) & 0.93(3.0) & 0.92(4.0) & 0.86(6.0) & 0.97(1.5)  \\ 
  ver & 0.68(5.5) & 0.78(4.0) & 0.68(5.5) & 0.84(2.0) & 0.82(3.0) & 0.89(1.0)  \\ 
  wal & 0.86(1.0) & 0.73(5.0) & 0.78(3.0) & 0.77(4.0) & 0.70(6.0) & 0.85(2.0)  \\ 
  wa1 & 0.77(6.0) & 0.81(4.0) & 0.79(5.0) & 0.83(3.0) & 0.84(2.0) & 0.90(1.0)  \\ 
  wa2 & 0.77(6.0) & 0.80(4.0) & 0.78(5.0) & 0.82(3.0) & 0.86(2.0) & 0.93(1.0)  \\ 
  win & 0.98(2.5) & 0.96(5.0) & 0.95(6.0) & 0.98(2.5) & 0.97(4.0) & 0.99(1.0)  \\ 
  yea & 0.82(3.5) & 0.79(5.0) & 0.89(1.0) & 0.88(2.0) & 0.72(6.0) & 0.82(3.5)  \\ 
  \hline
  avg & 0.88(3.13) & 0.83(4.61) & 0.87(3.63) & 0.89(2.84) & 0.85(4.46) & 0.91(2.33)  \\ 
 \end{tabular}
 }
 \caption{AUROC scores and ranks of algorithms using the second hyperparameter selection criterion.} 
 \label{tab:trainaucfull} 
\end{table}
\begin{table} 
 \center 
 \resizebox{\columnwidth}{!}{ 
 \begin{tabular}[h]{c | c c c c c c } 
  dataset & kNN & IForest & AE & VAE & GAN & fmGAN  \\ 
  \hline 
  aba & 0.93(1.0) & 0.83(4.0) & 0.89(2.0) & 0.87(3.0) & 0.74(6.0) & 0.77(5.0)  \\ 
  blo & 0.99(1.0) & 0.98(4.0) & 0.98(4.0) & 0.98(4.0) & 0.98(4.0) & 0.98(4.0)  \\ 
  brc & 0.93(4.0) & 0.98(1.0) & 0.95(3.0) & 0.96(2.0) & 0.86(6.0) & 0.87(5.0)  \\ 
  brt & 0.99(2.0) & 0.99(2.0) & 0.99(2.0) & 0.98(4.0) & 0.95(5.5) & 0.95(5.5)  \\ 
  car & 0.66(4.0) & 0.70(3.0) & 0.63(5.0) & 0.61(6.0) & 0.71(2.0) & 0.77(1.0)  \\ 
  eco & 0.93(3.0) & 0.87(4.0) & 0.95(1.5) & 0.95(1.5) & 0.80(5.0) & 0.79(6.0)  \\ 
  gla & 0.90(5.0) & 0.84(6.0) & 0.91(4.0) & 0.92(3.0) & 0.93(2.0) & 0.94(1.0)  \\ 
  hab & 0.97(1.0) & 0.94(2.5) & 0.86(4.0) & 0.94(2.5) & 0.85(5.0) & 0.83(6.0)  \\ 
  ion & 0.94(4.0) & 0.92(5.0) & 0.95(3.0) & 0.97(1.0) & 0.82(6.0) & 0.96(2.0)  \\ 
  iri & 0.97(4.5) & 0.92(6.0) & 0.97(4.5) & 0.98(3.0) & 1.00(1.5) & 1.00(1.5)  \\ 
  iso & 0.78(1.0) & 0.58(5.0) & 0.74(2.5) & 0.74(2.5) & 0.57(6.0) & 0.69(4.0)  \\ 
  let & 0.80(1.0) & 0.56(5.0) & 0.78(2.5) & 0.78(2.5) & 0.55(6.0) & 0.68(4.0)  \\ 
  lib & 0.87(1.5) & 0.74(6.0) & 0.82(4.0) & 0.83(3.0) & 0.76(5.0) & 0.87(1.5)  \\ 
  mad & 0.57(5.0) & 0.58(3.5) & 0.56(6.0) & 0.58(3.5) & 0.59(2.0) & 0.64(1.0)  \\ 
  mag & 0.95(1.0) & 0.86(4.0) & 0.92(2.5) & 0.92(2.5) & 0.78(6.0) & 0.83(5.0)  \\ 
  min & 0.75(5.0) & 0.85(4.0) & 0.73(6.0) & 0.91(1.5) & 0.87(3.0) & 0.91(1.5)  \\ 
  mul & 0.99(1.5) & 0.80(5.0) & 0.98(3.0) & 0.99(1.5) & 0.72(6.0) & 0.92(4.0)  \\ 
  mus & 0.97(3.0) & 0.60(6.0) & 0.98(2.0) & 0.99(1.0) & 0.68(5.0) & 0.80(4.0)  \\ 
  pag & 0.89(6.0) & 0.96(3.0) & 0.98(1.5) & 0.98(1.5) & 0.91(5.0) & 0.95(4.0)  \\ 
  par & 0.94(1.0) & 0.86(4.0) & 0.91(2.0) & 0.83(5.0) & 0.78(6.0) & 0.87(3.0)  \\ 
  pen & 1.00(1.0) & 0.91(4.0) & 0.98(2.5) & 0.98(2.5) & 0.81(6.0) & 0.82(5.0)  \\ 
  pim & 0.89(1.0) & 0.86(2.0) & 0.85(3.0) & 0.80(4.0) & 0.79(5.5) & 0.79(5.5)  \\ 
  son & 0.81(4.5) & 0.81(4.5) & 0.80(6.0) & 0.83(3.0) & 0.95(1.0) & 0.93(2.0)  \\ 
  spe & 0.89(2.5) & 0.86(6.0) & 0.90(1.0) & 0.88(4.0) & 0.87(5.0) & 0.89(2.5)  \\ 
  ssa & 0.97(1.0) & 0.92(5.0) & 0.95(2.5) & 0.95(2.5) & 0.87(6.0) & 0.93(4.0)  \\ 
  sse & 0.96(1.0) & 0.84(6.0) & 0.92(2.5) & 0.92(2.5) & 0.85(5.0) & 0.88(4.0)  \\ 
  ssh & 1.00(1.0) & 0.96(4.0) & 0.99(2.0) & 0.98(3.0) & 0.81(6.0) & 0.92(5.0)  \\ 
  sve & 0.83(3.0) & 0.79(5.0) & 0.84(2.0) & 0.86(1.0) & 0.74(6.0) & 0.81(4.0)  \\ 
  syn & 0.97(1.5) & 0.86(5.0) & 0.94(3.0) & 0.92(4.0) & 0.85(6.0) & 0.97(1.5)  \\ 
  ver & 0.76(2.0) & 0.78(1.0) & 0.68(6.0) & 0.75(3.0) & 0.74(4.5) & 0.74(4.5)  \\ 
  wal & 0.86(1.0) & 0.73(5.0) & 0.75(3.5) & 0.75(3.5) & 0.62(6.0) & 0.79(2.0)  \\ 
  wa1 & 0.77(5.5) & 0.81(1.5) & 0.79(3.5) & 0.79(3.5) & 0.77(5.5) & 0.81(1.5)  \\ 
  wa2 & 0.77(4.5) & 0.80(1.5) & 0.78(3.0) & 0.80(1.5) & 0.75(6.0) & 0.77(4.5)  \\ 
  win & 0.98(2.0) & 0.96(5.0) & 0.95(6.0) & 0.98(2.0) & 0.97(4.0) & 0.98(2.0)  \\ 
  yea & 0.82(3.0) & 0.79(4.0) & 0.86(2.0) & 0.87(1.0) & 0.70(6.0) & 0.76(5.0)  \\ 
  \hline
  avg & 0.88(2.57) & 0.83(4.07) & 0.86(3.24) & 0.86(2.73) & 0.76(4.9) & 0.80(3.49)  \\ 
 \end{tabular}
 }
 \caption{AUROC scores and ranks of algorithms using the third hyperparameter selection criterion, using top 5\% of samples.} 
 \label{tab:top5aucfull} 
\end{table}
\begin{table} 
 \center 
 \resizebox{\columnwidth}{!}{ 
 \begin{tabular}[h]{c | c c c c c c } 
  dataset & kNN & IForest & AE & VAE & GAN & fmGAN  \\ 
  \hline 
  aba & 0.93(1.0) & 0.82(4.0) & 0.86(2.0) & 0.84(3.0) & 0.67(6.0) & 0.73(5.0)  \\ 
  blo & 0.98(1.5) & 0.98(1.5) & 0.95(5.0) & 0.96(3.5) & 0.96(3.5) & 0.92(6.0)  \\ 
  brc & 0.93(3.0) & 0.98(1.0) & 0.92(4.0) & 0.94(2.0) & 0.82(5.0) & 0.80(6.0)  \\ 
  brt & 0.99(2.0) & 0.99(2.0) & 0.99(2.0) & 0.96(4.0) & 0.93(6.0) & 0.94(5.0)  \\ 
  car & 0.66(3.5) & 0.70(2.0) & 0.64(5.0) & 0.62(6.0) & 0.66(3.5) & 0.75(1.0)  \\ 
  eco & 0.93(1.5) & 0.88(3.5) & 0.88(3.5) & 0.93(1.5) & 0.86(5.0) & 0.83(6.0)  \\ 
  gla & 0.90(2.5) & 0.84(6.0) & 0.90(2.5) & 0.88(4.0) & 0.87(5.0) & 0.92(1.0)  \\ 
  hab & 0.97(1.0) & 0.94(2.0) & 0.86(4.0) & 0.91(3.0) & 0.82(5.0) & 0.79(6.0)  \\ 
  ion & 0.96(1.0) & 0.93(3.5) & 0.95(2.0) & 0.93(3.5) & 0.87(5.0) & 0.86(6.0)  \\ 
  iri & 0.96(1.0) & 0.92(5.5) & 0.94(3.5) & 0.94(3.5) & 0.95(2.0) & 0.92(5.5)  \\ 
  iso & 0.78(1.0) & 0.58(5.5) & 0.74(2.5) & 0.74(2.5) & 0.58(5.5) & 0.69(4.0)  \\ 
  let & 0.80(1.0) & 0.56(5.0) & 0.78(2.5) & 0.78(2.5) & 0.55(6.0) & 0.65(4.0)  \\ 
  lib & 0.87(1.5) & 0.74(6.0) & 0.82(4.0) & 0.83(3.0) & 0.80(5.0) & 0.87(1.5)  \\ 
  mad & 0.57(5.0) & 0.58(3.5) & 0.56(6.0) & 0.58(3.5) & 0.59(2.0) & 0.63(1.0)  \\ 
  mag & 0.95(1.0) & 0.86(4.0) & 0.90(2.0) & 0.89(3.0) & 0.77(6.0) & 0.79(5.0)  \\ 
  min & 0.75(5.0) & 0.84(2.0) & 0.72(6.0) & 0.92(1.0) & 0.83(3.0) & 0.81(4.0)  \\ 
  mul & 0.99(1.5) & 0.80(4.0) & 0.98(3.0) & 0.99(1.5) & 0.73(5.5) & 0.73(5.5)  \\ 
  mus & 0.97(3.0) & 0.58(6.0) & 0.98(1.5) & 0.98(1.5) & 0.65(5.0) & 0.69(4.0)  \\ 
  pag & 0.89(4.0) & 0.96(3.0) & 0.98(1.0) & 0.97(2.0) & 0.72(6.0) & 0.73(5.0)  \\ 
  par & 0.94(1.0) & 0.86(3.0) & 0.89(2.0) & 0.82(5.0) & 0.73(6.0) & 0.84(4.0)  \\ 
  pen & 1.00(1.0) & 0.90(4.0) & 0.97(2.5) & 0.97(2.5) & 0.76(5.0) & 0.72(6.0)  \\ 
  pim & 0.89(1.0) & 0.87(2.0) & 0.82(4.0) & 0.85(3.0) & 0.74(5.0) & 0.70(6.0)  \\ 
  son & 0.81(4.5) & 0.81(4.5) & 0.80(6.0) & 0.83(3.0) & 0.87(2.0) & 0.93(1.0)  \\ 
  spe & 0.89(3.0) & 0.86(6.0) & 0.91(1.0) & 0.90(2.0) & 0.87(4.5) & 0.87(4.5)  \\ 
  ssa & 0.97(1.0) & 0.92(3.0) & 0.91(4.0) & 0.93(2.0) & 0.84(6.0) & 0.89(5.0)  \\ 
  sse & 0.96(1.0) & 0.84(5.0) & 0.92(2.0) & 0.91(3.0) & 0.73(6.0) & 0.86(4.0)  \\ 
  ssh & 1.00(2.0) & 0.96(4.0) & 1.00(2.0) & 1.00(2.0) & 0.75(6.0) & 0.79(5.0)  \\ 
  sve & 0.83(1.0) & 0.79(4.0) & 0.82(2.0) & 0.78(5.0) & 0.68(6.0) & 0.80(3.0)  \\ 
  syn & 0.97(1.0) & 0.87(5.0) & 0.93(2.0) & 0.92(3.0) & 0.71(6.0) & 0.88(4.0)  \\ 
  ver & 0.68(5.0) & 0.78(1.0) & 0.69(4.0) & 0.71(3.0) & 0.73(2.0) & 0.65(6.0)  \\ 
  wal & 0.86(1.0) & 0.72(5.0) & 0.74(3.5) & 0.74(3.5) & 0.61(6.0) & 0.77(2.0)  \\ 
  wa1 & 0.77(5.0) & 0.80(1.0) & 0.79(3.0) & 0.79(3.0) & 0.72(6.0) & 0.79(3.0)  \\ 
  wa2 & 0.77(4.0) & 0.79(1.0) & 0.78(2.5) & 0.78(2.5) & 0.73(6.0) & 0.74(5.0)  \\ 
  win & 0.98(1.5) & 0.96(3.0) & 0.95(4.0) & 0.98(1.5) & 0.94(5.5) & 0.94(5.5)  \\ 
  yea & 0.82(1.0) & 0.79(2.0) & 0.78(3.0) & 0.74(4.0) & 0.68(6.0) & 0.69(5.0)  \\ 
  \hline
  avg & 0.89(2.14) & 0.83(3.53) & 0.87(3.13) & 0.88(2.93) & 0.80(4.97) & 0.85(4.3)  \\ 
 \end{tabular}
 }
 \caption{AUROC scores and ranks of algorithms using the third hyperparameter selection criterion, using top 1\% of samples.} 
 \label{tab:top1aucfull} 
\end{table}

\end{document}